\documentclass[letterpaper, 10 pt, conference]{IEEEtran}
\IEEEoverridecommandlockouts
\newcommand\copyrighttext{%
	\footnotesize This work has been submitted to the IEEE for possible publication. Copyright may be transferred without notice, after which this version may no longer be accessible.
}
\newcommand\copyrightnotice{%
	\begin{tikzpicture}[remember picture,overlay]
	\node[anchor=south,yshift=10pt, xshift=10pt] at (current page.south) {\fbox{\parbox{\dimexpr\textwidth-\fboxsep-\fboxrule\relax}{\copyrighttext}}};
	\end{tikzpicture}%
}

\usepackage{printlen}

\usepackage[T1]{fontenc}
\usepackage{amsmath,amsfonts}
\usepackage{cite}
\usepackage{stfloats}
\usepackage{graphicx}
\usepackage[caption=false,font=scriptsize]{subfig}
\usepackage{url}
\usepackage{hyperref}

\usepackage[per-mode = fraction]{siunitx}
\usepackage{acronym}
\usepackage[none]{hyphenat}

\usepackage{bbm} 
\usepackage{blindtext}
\usepackage[]{glossaries}

\usepackage{booktabs}
\usepackage{tabularx}
\usepackage{multirow}
\usepackage{pifont}

\usepackage[linesnumbered]{algorithm2e}
\RestyleAlgo{ruled}
\DontPrintSemicolon 

\usepackage{tikz, pgfplots, pgfplotstable}
\usetikzlibrary{arrows, arrows.meta, patterns}
\usepgfplotslibrary{statistics}
\pgfplotsset{compat=1.18}
\usetikzlibrary{external}
\tikzset{external/only named=true}

\usepackage{xcolor}
\definecolor{Black}{HTML}{000000}
\definecolor{Blue}{HTML}{0065bd}
\definecolor{Bluelight}{HTML}{D6E8F7}
\definecolor{Bluestrong}{HTML}{003359}
\definecolor{Red}{HTML}{8C000F}
\definecolor{Orange}{HTML}{E37222}
\definecolor{OrangePP}{HTML}{E97132}
\definecolor{Green}{HTML}{A2AD00}
\definecolor{GreenCR}{HTML}{008000}
\definecolor{LightGray}{HTML}{e7e7e7}
\definecolor{Gray}{HTML}{7f7f7f}
\definecolor{Gray-opac}{HTML}{d8d8d8}
\definecolor{GreenTraj}{HTML}{c3e57c}
\definecolor{OrangeTraj}{HTML}{fdc675}

\graphicspath{{./figures/}}

\usepackage{textcomp} %
\DeclareRobustCommand{\reg}{\textsuperscript{\textregistered}}

\title{\LARGE \bf
Towards Safe Autonomous Driving: A Real-Time Motion Planning Algorithm on Embedded Hardware
}

\newif\iffinal
\finaltrue

\newcommand{\authorsFinal}{
    \author{Korbinian Moller, Glenn Johannes Tungka, Lucas Jürgens, Johannes Betz%
    \thanks{K. Moller, G. Tungka, L. Jürgens and J. Betz are with the Professorship of Autonomous Vehicle Systems, TUM School of Engineering and Design, Technical University of Munich, 85748 Garching, Germany; Munich Institute of Robotics and Machine Intelligence (MIRMI).}%
    }
}

\newcommand{\authorsBlind}{%
  \author{Anonymous Author(s)%
  \thanks{This work has been submitted to IV 2026 for peer review. Affiliations and links to the open-source code are omitted for double-blind review.}}
}

\iffinal
  \authorsFinal
\else
  \authorsBlind
\fi

\newacronym{ads}{ADS}{Autonomous Driving System}%
\newacronym{av}{AV}{Autonomous Vehicle}%
\newacronym{ad}{AD}{Autonomous Driving}%
\newacronym{ev}{EV}{Ego Vehicle}%
\newacronym{hpc}{HPC}{High-Performance Computer}%
\newacronym{rtos}{RTOS}{Real-Time Operating System}%
\newacronym{ov}{OV}{Online Verification}%
\newacronym{ros}{ROS}{Robot Operating System}%
\newacronym{ros2}{ROS~2}{Robot Operating System 2}%
\newacronym{dds}{DDS}{Data Distribution Service}%

\begin{document}
\bstctlcite{BSTcontrol}

\maketitle
\copyrightnotice


\begin{abstract}
Ensuring the functional safety of \acrfullpl{av} requires motion planning modules that not only operate within strict real-time constraints but also maintain controllability in case of system faults. Existing safeguarding concepts, such as \acrfull{ov}, provide safety layers that detect infeasible planning outputs. However, they lack an active mechanism to ensure safe operation in the event that the main planner fails.
This paper presents a first step toward an active safety extension for fail-operational \gls{ad}. We deploy a lightweight sampling-based trajectory planner on an automotive-grade, embedded platform running a \acrfull{rtos}. The planner continuously computes trajectories under constrained computational resources, forming the foundation for future emergency planning architectures. Experimental results demonstrate deterministic timing behavior with bounded latency and minimal jitter, validating the feasibility of trajectory planning on safety-certifiable hardware. The study highlights both the potential and the remaining challenges of integrating active fallback mechanisms as an integral part of next-generation safeguarding frameworks. The code is available at: \url{https://github.com/TUM-AVS/real-time-motion-planning}
\end{abstract}



\section{Introduction}
\label{sec:introduction}

Over the last few years, \glspl{av} have gained significant attention due to their potential to enhance traffic efficiency and accessibility~\cite{Nastjuk2020}. Furthermore, by reducing human error, the predominant cause of traffic accidents, \glspl{av} promise to significantly lower crash rates.

Despite these advantages, the deployment of \glspl{av} in complex environments remains a challenge. Motion planning, in particular, is a safety-critical component that must operate under strict timing and memory constraints~\cite{Tezerjani2025}. If trajectory computation exceeds its deadline or produces infeasible outputs, vehicle controllability can no longer be guaranteed. To mitigate this risk, several research efforts have focused on \acrfull{ov} and runtime safeguarding, where an independent module continuously checks the feasibility and timing of planning outputs~\cite{Stahl2020OnlineVerification, Moller2025Safeguarding}. These approaches enable ASIL decomposition in accordance with ISO 26262~\cite{ISO26262}, allowing a safety layer to supervise complex, nondeterministic planners. However, these mechanisms often remain passive.

For real fail-operational safety, these safeguarding concepts must be complemented by an active safety component capable of providing a feasible fallback solution~\cite{Moller2025Safeguarding}. A promising direction is to execute a lightweight trajectory planner on an independent, real-time platform, called Safety Island~\cite {ARM2025safetyisland}. This planner should continuously generate fallback trajectories that the control stack uses in case of a failure. Compared to a fully redundant system on a second high-performance unit, this approach offers a cost-efficient and resource-aware path for active safety systems.

\begin{figure}[!t]
    \centering
    \includegraphics[width=0.49\textwidth, trim={0cm 0cm 0cm 0cm},clip]{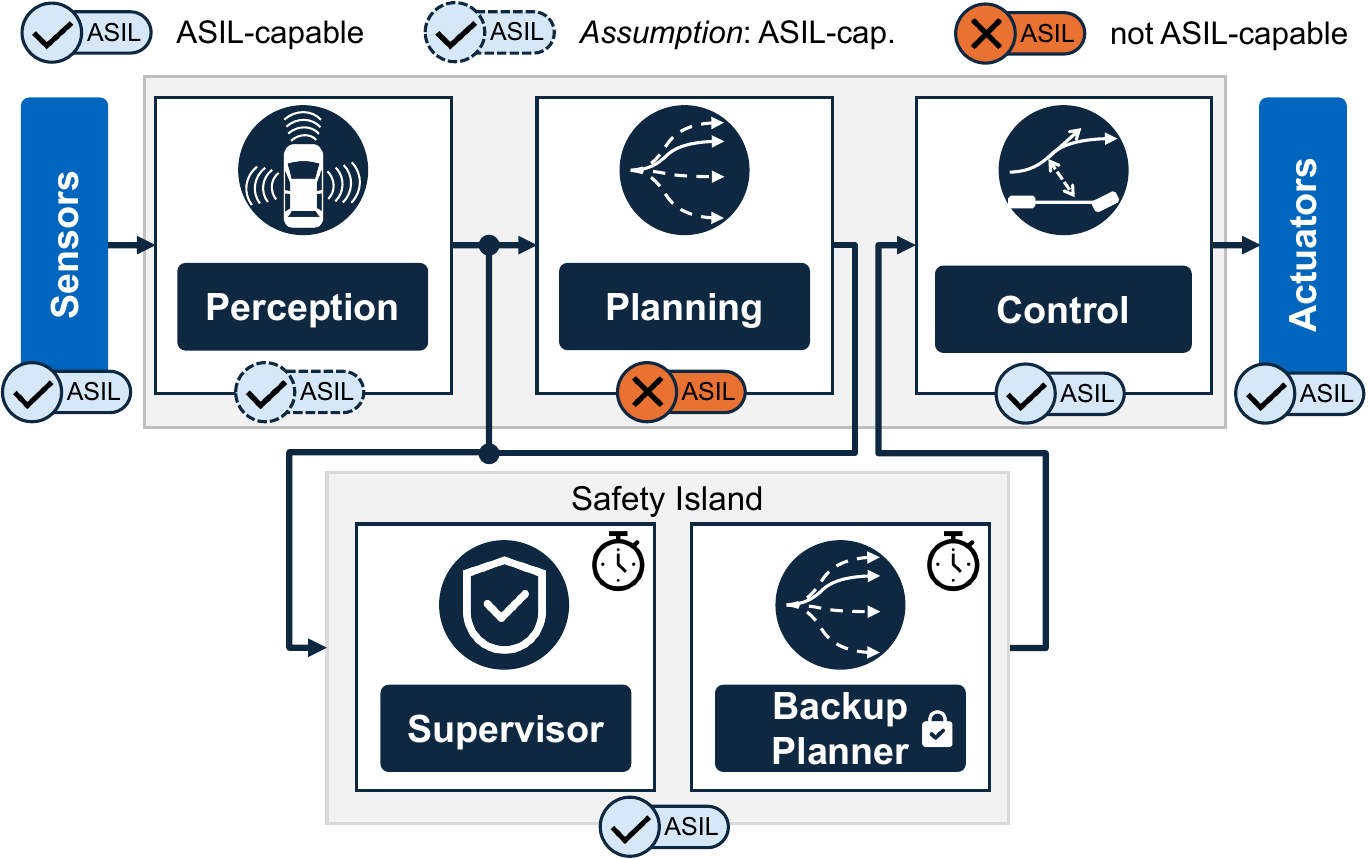}
    \vspace{-6mm}
    \caption{System architecture illustrating the integration of a safety island that hosts both a Supervisor and an embedded planner. The concept enables safety monitoring and fallback trajectory generation independently of the main software stack, forming the foundation for fail-operational \gls{ad}.}
    \label{fig:futureimpact}
\end{figure}

The long-term objective of this research is to integrate such an embedded backup planner as an active extension to an existing safeguarding framework. While the envisioned system will eventually run in parallel with the main planning module, this work represents an important first step toward that goal. We demonstrate the feasibility of deploying a sampling-based trajectory planner~\cite{Trauth2024frenetix} on an automotive-grade, embedded platform running a \gls{rtos}. The focus lies on achieving deterministic timing and reliable computation under constrained resources, thereby quantifying both the potential and the limitations of real-time planning on safety-certifiable hardware. An overview of the envisioned architecture is shown in \autoref{fig:futureimpact}. The main contributions of this paper are as follows:

\begin{itemize}
\item A safety-oriented planning concept that can extend runtime safeguarding by introducing a fallback mechanism for fail-operational \gls{ad}.
\item An embedded real-time implementation of a sampling-based trajectory planner optimized for deterministic execution under limited resources.
\item An empirical evaluation of latency, jitter, and CPU utilization, providing insight into the feasibility and challenges of deploying active safety logic on constrained hardware.
\end{itemize}

\section{Related Work}
\label{sec:relatedwork}

In the context of \gls{ad}, motion planning is not only a question of correctness but also of timing. A motion planner must produce safe and feasible trajectories, yet the computation time itself becomes a constraint: results are only useful if they are available before a strict deadline. This requirement is a major concern of real-time computing, where the decisive criterion is not only high throughput or performance, but the guarantee that a task completes reliably within its allocated time window~\cite{Buttazzo2024}. For \glspl{av}, this means that planning algorithms must balance responsiveness and fidelity with deterministic execution, even when running under constrained resources. In recent years, several planning concepts have been developed~\cite{Teng2023}. 
These methods can be categorized into optimization-, graph-, sampling-, and machine learning-based algorithms~\cite{Hu2025Survey}. Every class offers specific trade-offs in terms of runtime or predictability properties.

Optimization-based methods formulate motion planning as a constrained optimization problem, in which a cost function is minimized subject to vehicle dynamics, actuator limitations, and environmental constraints~\cite{Li2023, Arjmandzadeh2024}. Approaches such as Quadratic Programming and Model Predictive Control (MPC)~\cite{Liu2017MPCPathPlanning} optimize trajectories by calculating future system states and guaranteeing feasibility at each step. These methods yield smooth and consistent motion profiles and allow the explicit incorporation of safety-relevant constraints. However, their computational demand is considerable, especially in dynamic environments, where repeated online optimization becomes a bottleneck. Direct optimal-control formulations~\cite{Luo2020OptimizationMotionPlanning} further improve accuracy but still face challenges regarding runtime, memory consumption, and predictable execution.

Graph-based methods discretize the planning space into nodes and edges, enabling the use of established search algorithms such as A*~\cite{Hart1968AStar}. 
They are especially effective in structured environments~\cite{Dolgov2010HybridAStar}, where a sufficiently fine grid guarantees that a drivable path can be found if one exists. Nevertheless, high spatial resolution quickly increases computational requirements, and the resulting paths often need smoothing before they can be executed by a vehicle~\cite{Huang2023}. Moreover, classic graph-based planners operate in the spatial domain and therefore do not handle dynamic obstacles well. To incorporate temporal information, extensions such as spatiotemporal graphs~\cite{Shu2025, Rowold2022} augment the graph structure with velocity and timing profiles, yielding full spatiotemporal trajectories at the cost of additional computational effort.

Sampling-based methods compute a finite set of trajectory candidates by sampling in the action or control space around the current \gls{ev} state~\cite{Werling2010, Trauth2024frenetix, Kim2025}. Each candidate is evaluated with respect to feasibility constraints, such as curvature or acceleration and ranked using cost functions that capture safety, comfort, and efficiency~\cite{Huang2023}. The computational load can be adjusted by varying the number of samples, making these methods particularly suitable for real-time applications where predictable runtime is essential. However, planning quality depends on sampling density and coverage~\cite{Trauth2024frenetix}: insufficient sampling may miss feasible or desirable maneuvers, while excessive sampling increases runtime. Despite these limitations, sampling-based planners remain a widely used approach between expressiveness, robustness, and computational tractability.

\section{Methodology}
\label{sec:method}

\begin{figure*}[!t]
    \centering
    \includegraphics[width=0.8\linewidth]{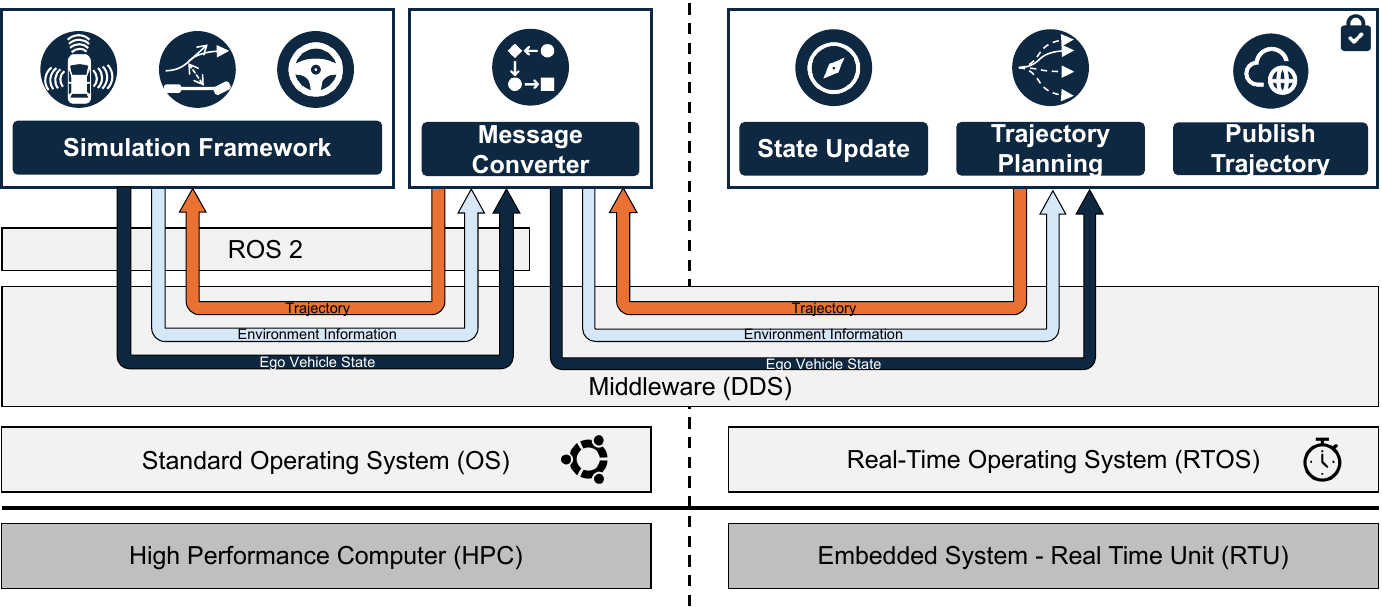}
    \vspace{-3mm}
    \caption{Overview of the proposed communication architecture between the \gls{hpc} and the Embedded System. The \gls{hpc} runs a \gls{ros2} simulation, while the \gls{rtos} executes our motion planner. A lightweight DDS bridge enables data exchange across both platforms.}
    \label{fig:methodology}%
\end{figure*}

While numerous motion planning frameworks have been proposed to achieve real-time capability in \gls{ad}~\cite{Tezerjani2025, Cheng2022, Xu2012-RealTimePlanner}, most implementations operate on \glspl{hpc} using general-purpose operating systems. These environments provide large computational resources but lack deterministic scheduling and guaranteed timing behavior. This work, therefore, investigates the deployment of a motion planner on an embedded platform running an \gls{rtos}. The \gls{rtos} provides preemptive scheduling of tasks, bounded interrupt latency, and a low-footprint execution environment, enabling the implementation of applications satisfying hard real-time requirements under constrained compute power. The presented system represents a fully functional planning stack, operating as the primary decision-making component within an \gls{ad} architecture. It interacts with perception, localization, and control subsystems through communication interfaces. The system overview is shown in \autoref{fig:methodology}.

\subsection{Prerequisites and Assumptions}

Our motion planner assumes input information is provided by the \gls{ad} stack. At each planning step \( t \), the planner requires the current \gls{ev} state $\mathbf{x}_t$ in cartesian coordinates,
\begin{equation}
\mathbf{x}_t = (x_t, y_t, v_t, a_t, \theta_t)^\top,
\label{eq:state}
\end{equation}
where \( (x_t, y_t) \) denote the \gls{ev} position in the world frame, \( v_t \), \( a_t \) and \( \theta_t \) represent the current velocity, acceleration and orientation, respectively. The \gls{ev} state is assumed to be estimated by a localization module.

Our motion planner further needs a geometrically valid reference path \( \Gamma \), defined as a continuously differentiable curve
\begin{equation}
\Gamma = \{(x_r(s), y_r(s), \theta_r(s)) \mid s \in [0, S]\}, 
\end{equation}
which provides a reference for local trajectory planning. To maintain a bijective mapping between cartesian and curvilinear coordinates, \( \Gamma \) has to be non-self-intersecting and to exhibit bounded curvature \( |\kappa_r(s)| < \kappa_{\max} \) for all \( s \in [0, S] \). This ensures that every cartesian point in the projection domain along $\Gamma$ can be transformed to a unique longitudinal coordinate $s$ and lateral coordinate $d$.

In addition to $\Gamma$, the environment contains a finite set of objects
\begin{equation}
\mathcal{O}_t = \{o_t^1, o_t^2, \dots, o_t^M\},
\end{equation}
where each object \( o_t^j \) is characterized by a predicted trajectory over a finite horizon, provided by an upstream perception and prediction module.

The planner’s output at each cycle is a single trajectory \( \xi_t \), represented as a discrete sequence of states over a finite horizon \( \tau \) with uniform time increment \( \Delta t \):
\begin{equation}
\xi_t = \{\mathbf{x}_0, \mathbf{x}_1, \dots, \mathbf{x}_N\}.
\end{equation}
Each state of \( \xi_t \), as described in \autoref{eq:state}, corresponds to the planned vehicle motion at time \( t + i \Delta t \). The trajectory serves as input for a downstream control module that executes the trajectory and compensates for external disturbances.

\subsection{Communication Interface}

Current robotics or \gls{ad} applications such as Autoware\footnote{\url{https://autoware.org/}} or high-performance autonomous racing stacks~\cite{Betz2023} often rely on \acrshort{ros2} as middleware due to its modular architecture and broad ecosystem support. Despite its flexibility, \acrshort{ros2} is not designed for hard real-time operation, as its message-passing concept introduces nondeterministic scheduling and dynamic memory allocation overheads~\cite{Teper2025}. Due to the high coupling of \acrshort{ros2} to Linux, including its dependencies, e.g., Python or a file system, \acrshort{ros2} applications cannot be ported directly to \gls{rtos} like Zephyr\footnote{\url{https://www.zephyrproject.org/}}. However, multiple \gls{dds} clients or micro-\acrshort{ros} implementations are available to enable data exchange between different systems. In this work, communication between the high-performance domain and the embedded planner is realized through a minimal \gls{dds} bridge. The high-performance node, running the \acrshort{ros}-based software stack, publishes the \gls{ev}-state \(\mathbf{x}_t \), reference path \( \Gamma \), and surrounding obstacles \( \mathcal{O}_t \), which are translated into DDS messages and sent to the \gls{rtos} domain~\cite{Moller2025Safeguarding}. The embedded planner subscribes to these \gls{dds} topics, performs motion planning, and returns the resulting trajectory through the same interface. This design maintains full compatibility with \acrshort{ros2}-based systems without the need to run a complete \acrshort{ros2} runtime on the embedded platform.

\subsection{Execution Workflow on the Embedded Platform}

\begin{figure}[!t]
    \centering
    \includegraphics[width=0.82\linewidth]{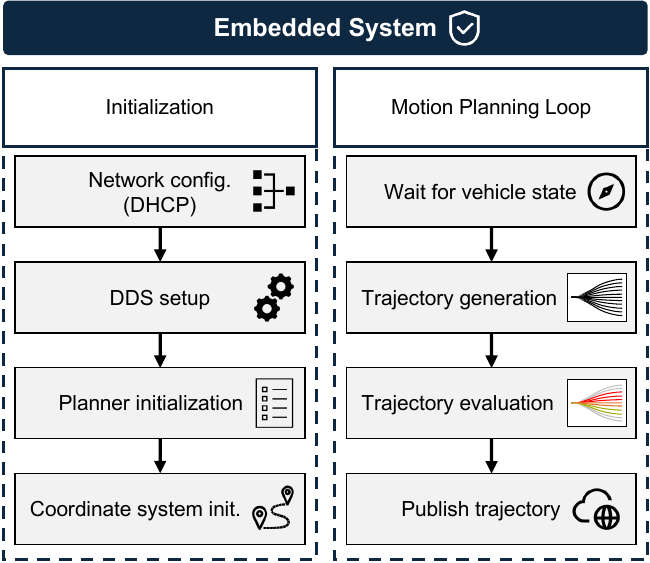}
    \caption{Overview of the planning workflow on the embedded platform. The initialization phase (left) is executed once at startup, followed by the cyclic motion planning loop (right).}
    \label{fig:workflow_overview}
\end{figure}

Our motion planner is executed as a standalone module on the embedded platform and follows a cyclic workflow, as shown in \autoref{fig:workflow_overview}.

Each planning cycle begins with the initialization stage, where the system configures its network connection, starts the DDS client and initializes the motion planner. It also processes the reference path and creates the curvilinear coordinate system. This setup is performed once at startup and remains static during runtime.

During the online phase, the planner iteratively performs four steps. First, the current \gls{ev} state \( \mathbf{x}_t \) is received via the \gls{dds} interface and stored in shared memory accessible to the planning thread. Second, the planner generates candidate trajectories \( \xi \in \mathcal{T} \) based on predefined sampling parameters. Third, each trajectory is evaluated according to feasibility and basic cost metrics, as detailed in the following section. Finally, the selected trajectory \( \xi^{\star} \) is serialized and transmitted back to the \gls{hpc}. The cycle repeats continuously until the scenario goal is reached or the execution is terminated.

\subsection{Trajectory Generation and Evaluation}
\label{subsec:traj_generation_and_evaluation}
The trajectory generation process follows a sampling-based approach formulated in the curvilinear frame \((s,d)\), which expresses vehicle motion relative to the global reference path $\Gamma$~\cite{Werling2010}. A candidate trajectory \(\xi:[0,\tau]\!\to\!\mathbb{R}^m\) is defined by the mapping
\begin{equation}
\xi = \mathcal{P}\bigl(\mathbf{x}(0), g(\tau), \Gamma\bigr),
\end{equation}
where \(g(\tau)\) represents the sampled goal state at the end of the planning horizon $\tau$.  

Following the formulation in~\cite{Werling2010, Trauth2024frenetix}, longitudinal movement is represented by a quartic polynomial
\begin{equation}
s(t) = a_0 + a_1 t + a_2 t^2 + a_3 t^3 + a_4 t^4,
\end{equation}
and the lateral motion by a quintic polynomial
\begin{equation}
d(t) = b_0 + b_1 t + b_2 t^2 + b_3 t^3 + b_4 t^4 + b_5 t^5.
\end{equation}
The polynomial coefficients are analytically calculated from the boundary conditions created by the initial state \(\mathbf{x}(0)\) and \(g(\tau)\). For the lateral component, the goal velocity and acceleration are set to zero to ensure that the \gls{ev} aligns with \( \Gamma \) at the end of each trajectory sample~\cite{Werling2010}. This formulation minimizes jerk while maintaining low computational complexity.

Our planner generates a discrete set of goal states \(g_k(\tau)\) by sampling time, lateral offset, and goal velocity, following the approach described in~\cite{Trauth2024frenetix}. The sampling ranges are defined as
\begin{equation}
\tau \in [\tau_{\min}, \tau_{\max}],\,
d_\tau \in [d_{\min}, d_{\max}],\,
v_\tau \in [v_{\min}, v_{\max}],
\end{equation}
where each set \((\tau, d_\tau, v_\tau)\) yields one candidate trajectory $\xi$. By combining \(m\) longitudinal and \(n\) lateral samples, a set of \(m\times n\) candidate trajectories is created. Each trajectory is transformed back into cartesian coordinates using the curvilinear projection along \(\Gamma\)~\cite{Werling2010}.

All trajectory candidates are then evaluated for kinematic feasibility according to the vehicle’s physical limits. The longitudinal acceleration must satisfy
\begin{equation}
a_{\min} \leq a(t) \leq a_{\max}(t), \quad \forall t \in [t_0, \tau],
\end{equation}
and the curvature \(\kappa(t)\) must remain below the limit created by the maximum steering angle \(\delta_{\max}\) and wheelbase \(L\):
\begin{equation}
|\kappa(t)| \leq \frac{\tan(\delta_{\max})}{L}, \quad \forall t \in [t_0, \tau].
\end{equation}
Additional constraints apply to the rate of curvature change and the yaw rate:
\begin{align}
|\dot{\kappa}(t)| &\leq \dot{\kappa}_{\max}, \quad \forall t \in [t_0, \tau], \\
|\dot{\psi}(t)| &\leq \kappa_{\max} \cdot v(t), \quad \forall t \in [t_0, \tau].
\end{align}
Trajectories that violate any of these limits are discarded before being evaluated for cost, thereby reducing runtime.

The feasible trajectories are finally ranked according to a cost function that combines multiple comfort and safety criteria~\cite{Trauth2024frenetix}. Representative examples include the lateral deviation from $\Gamma$,
\begin{equation}
J_{\text{ref}}(\xi) = \sum_{i=0}^{N} \bigl(d_i^\perp\bigr)^2,
\end{equation}
the velocity offset with respect to a desired speed \(v_{\text{des}}\),
\begin{equation}
J_{\text{vel}}(\xi) = \sum_{i=0}^{N} \bigl|v_i - v_{\text{des}}\bigr|^p,
\end{equation}
and acceleration-based measures,
\begin{equation}
J_{\text{lat}}(\xi) = \int_{t_0}^{\tau} a_{\text{lat}}(t)^2\,\mathrm{d}t, \quad
J_{\text{lon}}(\xi) = \int_{t_0}^{\tau} a_{\text{lon}}(t)^2\,\mathrm{d}t.
\end{equation}
The total trajectory cost is calculated as a weighted sum
\begin{equation}
J_{\text{sum}}(\xi) = \sum_{i=1}^{n} \omega_i \, J_i(\xi),
\end{equation}
where \(\omega_i\) are the corresponding weighting factors. The optimal trajectory is selected as
\begin{equation}
\xi^* = \arg\min_{\xi \in \mathcal{T}_t^{\text{feas}}} (J_{\text{sum}}(\xi)).
\label{eq:total_costs}
\end{equation}

Executing this planning algorithm on Zephyr~\gls{rtos} with limited hardware resources introduces additional challenges that influence the implementation. Zephyr provides a lightweight runtime optimized for deterministic task scheduling, but it lacks full support for the C++ standard library. Consequently, several libraries commonly used in motion planning algorithms, such as Boost, are unavailable. Geometric operations required for feasibility checks, including polygon intersection and point-in-polygon tests, are therefore substituted using the lightweight Clipper2 library.

\section{Results \& Discussion}
\label{sec:results}

The following section presents the experimental evaluation of our proposed real-time motion planner, which is executed on the embedded platform under the Zephyr~\gls{rtos}. The objective of this study is to evaluate the computational performance of our planning algorithm under constrained hardware conditions and to derive insights for future extensions toward fail-operational planning architectures.  

The following evaluation focuses on runtime characteristics, e.g., latency, jitter, and CPU utilization. In addition, qualitative analyses are conducted to verify that the planning process preserves the expected planning behavior compared to the reference implementation~\cite{Trauth2024frenetix}.  

\subsection{Experimental Setup}

All experiments were conducted on an automotive-grade embedded platform based on the NXP\reg~S32Z2 real-time processor. The system uses an Arm\reg~Cortex-R52 core running at \SI{800}{\mega\hertz} under Zephyr~\gls{rtos} (v3.5). In our evaluation setup, the \gls{hpc} corresponds to a desktop system with \num{16} logical processors, \SI{32}{\giga\byte} RAM, and a base frequency of \SI{2.2}{\giga\hertz} running Ubuntu~22.04. The \gls{hpc} executes an adapted CommonRoad environment~\cite{CommonRoad}, enabling a reproducible \acrshort{ros2}-based evaluation of our planning framework. In all experiments, each trajectory was parameterized over a fixed horizon of 30 time steps with a resolution of $\Delta t = \SI{0.1}{\second}$. To isolate baseline timing behavior, no obstacles were considered during this evaluation. Each trajectory batch was visualized and checked for consistency.

\subsection{Qualitative Evaluation}

\begin{figure}[ht!]
\centering
\fbox{\includegraphics[width=7cm, trim={7cm 2.5cm 2cm 6.0cm},clip]{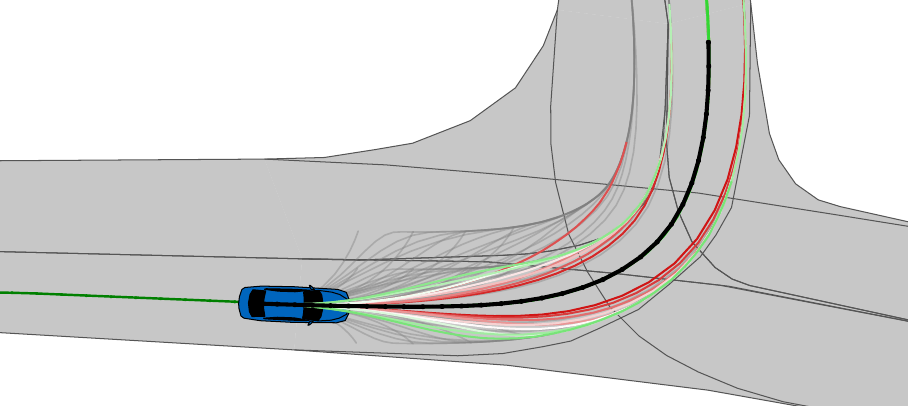}}
\caption{Qualitative simulation result in the CommonRoad environment, showing a left-turn scenario at an unsignalized intersection. The generated candidate trajectories follow the reference path, illustrating how the planner maintains smooth motion during the intended turning maneuver.}
\label{fig:qualitative_evaluation_simulation}
\end{figure}

For qualitative validation, we executed a standard CommonRoad~\cite{CommonRoad} scenario, in which the \gls{ev} must perform a left turn at an unsignalized intersection, as shown in \autoref{fig:qualitative_evaluation_simulation}. This scenario enables a thorough analysis of feasibility and cost calculation mechanisms, as well as verification of the cartesian-to-curvilinear projection. The generated trajectories are aligned with the lane boundaries, and both feasibility and cost evaluation yield consistent results within the CommonRoad simulation framework.

\begin{figure}[ht!]
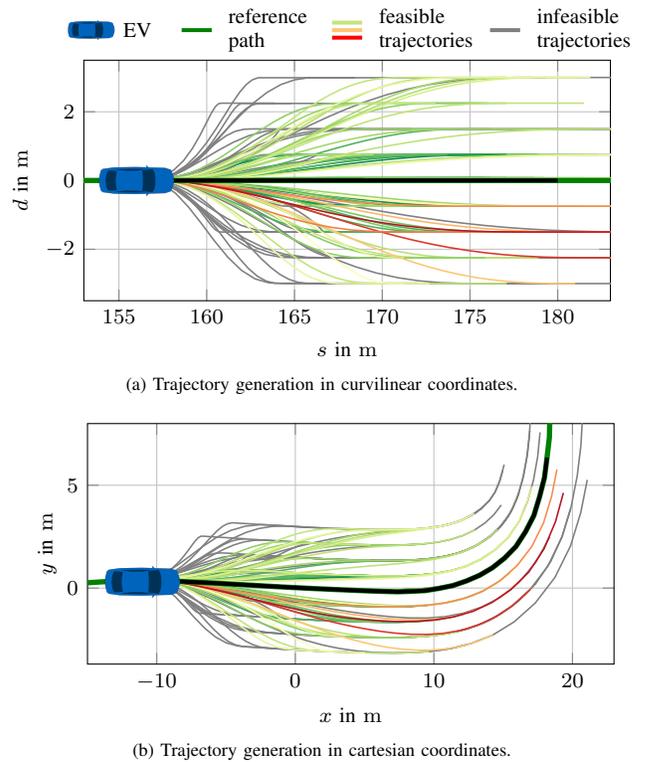

    \centering
    \input{figures/trajectory_colors}
    \subfloat[][\scriptsize{Trajectory generation in curvilinear coordinates.}]{\label{fig:trajectories_s_d_cost} %
    \tikzsetnextfilename{trajectories_s_d_cost}%
    \input{figures/trajectories_s_d_cost}%
} \\
    \subfloat[][\scriptsize{Trajectory generation in cartesian coordinates.}]{\label{fig:trajectories_x_y_cost} %
    \tikzsetnextfilename{trajectories_x_y_cost}%
    \input{figures/trajectories_x_y_cost}%
} \\    
    \caption{Generation and evaluation of trajectory samples by the planner along $\Gamma$. Valid trajectories are color-coded based on their cost, ranging from green (low cost) to red (high cost). The optimal trajectory is shown in black.}
    \label{fig:qualitative_evaluation_cart_curv}
\end{figure}

\autoref{fig:qualitative_evaluation_cart_curv} furthermore details the set of generated trajectory candidates, shown in both curvilinear (Figure~\ref{fig:trajectories_s_d_cost}) and cartesian (Figure~\ref{fig:trajectories_x_y_cost}) coordinates.
Each line represents one trajectory candidate $\xi$, where infeasible trajectories are drawn in gray and feasible ones are color-coded according to their total cost, ranging from green (low cost) to red (high cost). The optimal trajectory $\xi^{\star}$ is selected based on the overall cost function (\autoref{eq:total_costs}) and is shown in black. 

As introduced in Section~\ref{subsec:traj_generation_and_evaluation}, the sampling parameters directly influence the generated trajectory tree. Lateral offsets correspond to different displacements, while variations in goal velocity and time horizon affect the goal states in both position and timing. \autoref{fig:qualitative_evaluation_cart_curv} visualizes how these sampling parameters influence the shape and density of all generated trajectories $\mathcal{T}$. It also provides a representation of the planner’s evaluation logic, allowing for validation of its feasibility and cost calculation pipeline. Trajectories with high lateral offset from $\Gamma$ or large longitudinal acceleration are penalized through corresponding cost terms. Likewise, deviations from a desired cruise speed increase the total cost. No parameter tuning was done for this example; all weighting factors were selected manually, and the desired velocity was fixed to \SI{7.0}{\meter\per\second} for illustrative purposes only. While this visualization shows the evaluation of all trajectory candidates, a real-world deployment would only publish the optimal trajectory to the controller.

\subsection{Runtime and Latency Evaluation}

\begin{figure}[!ht]
    \centering
    \tikzsetnextfilename{latency-board-vs-hpc}%
    \input{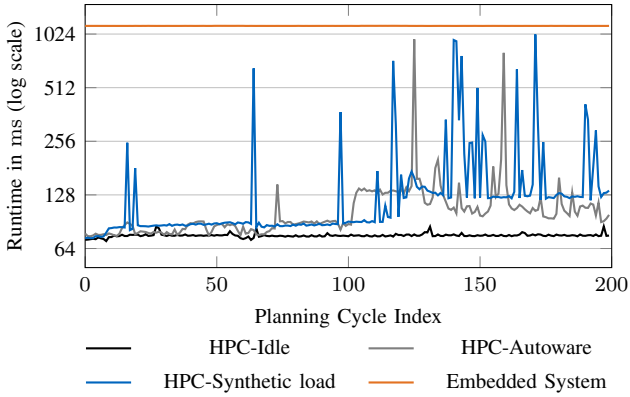}%

    \vspace{-0.4cm}
    \caption{Comparison of runtime behavior between the embedded RTOS-based platform and the HPC under varying computational load. The embedded board exhibits stable timing behavior with minimal jitter.}
    \label{fig:runtime_arm_vs_hpc}%
\end{figure}

To analyze the temporal behavior of our embedded planner, three metrics are evaluated: the average runtime \(T_{\text{avg}}\), the worst-case runtime \(T_{\text{max}}\), and the jitter \(J\). These are defined in the following equations, where \(T_i\) denotes the runtime of the \(i\)-th planning cycle and \(n\) the total number of iterations.
\begin{align}
T_{\text{avg}} &= \frac{1}{n} \sum_{i=1}^{n} T_i \\[1mm]
T_{\text{max}} &= \max(T_i) \\[2mm]
J &= \max \left( \left| T_i - T_{\text{avg}} \right| \right)
\end{align}
To compare the latency characteristics of the embedded platform with those of a regular \gls{hpc}, both systems were configured to execute the same planning task, consisting of 64 trajectory samples per cycle. On the \gls{hpc}, three configurations were tested: an idle setup without background load, a setup with the Autoware stack running concurrently, and a setup with synthetic load introduced on 8 of 16 cores.~\autoref{fig:runtime_arm_vs_hpc} illustrates the measured runtimes of consecutive planning cycles for both systems. 

As expected, the \gls{hpc} achieves lower average runtimes due to its higher processing power. However, under concurrent execution, significant temporal variability can be observed. When the background system load increases, the runtime distribution widens, and latency peaks occur as a consequence of process preemption and interference within the Linux scheduler. In contrast, our embedded platform running the \gls{rtos} maintains stable execution with a mean runtime of approximately \SI{1138}{\milli\second} and negligible jitter across all measurements.  

This comparison highlights a fundamental difference between high-performance and real-time systems. While Linux-based environments can, in principle, achieve real-time behavior through kernel-level tuning, adapted scheduling, and CPU pinning, such configurations require high engineering effort and are sensitive to kernel version, driver behavior, and system load. Ensuring deterministic behavior in such environments requires careful selection of the scheduler, strict control of thread priorities, and pre-allocating memory pages to prevent blocking operations. Even then, effects such as cache pollution or non-preemptive kernel modules (e.g., GPU drivers) can compromise timing predictability.  

In contrast, our Zephyr-based setup provides a deterministic execution environment by design. The RTOS enables bounded interrupt latency, fixed-priority scheduling, and complete control over memory allocation. Consequently, the runtime characteristics remain unaffected by external processes or concurrent tasks, making it a more reliable platform for safety-critical applications. From a system design perspective, this determinism enables a robust separation of safety domains. A complex AV software stack may operate on the \gls{hpc}, while the \gls{rtos} executes on the embedded platform. This architectural separation offers a path toward ASIL-compliant implementations, where the planner serves as a safety function operating independently of the main software stack.

To further analyze runtime behavior, the planner was executed with increasing numbers of trajectory samples (16, 32, 64, and 100). For 16 samples, the total runtime is around \SI{287}{\milli\second}, increasing to approximately \SI{1.78}{\second} for 100 samples. Despite the limited computational resources, the board maintains a low jitter across all configurations, confirming deterministic execution. At lower sample counts, the system operates within real-time bounds, sustaining a planning frequency of roughly \SI{2}{\hertz}.

This behavior shows the trade-off between trajectory diversity and computation time. Increasing the number of samples enhances coverage of the solution space and may yield smoother or safer trajectories, but comes at the cost of increased computation time. Table~\ref{tab:benchmark_ms_comparison} summarizes the results for 16, 32, 64, and 100 trajectories.

\begin{table}[!ht]
\centering
\caption{Timing metrics for different numbers of trajectory samples on the Embedded System.}
\label{tab:benchmark_ms_comparison}
\begin{tabular}{lcccc}
\toprule
\multirow{2}{*}[-0.5mm]{\textbf{Metric}} & \multicolumn{4}{c}{\textbf{Trajectories}}\\
\cmidrule(lr){2-5}
 & \textbf{16} & \textbf{32} & \textbf{64} & \textbf{100} \\
\midrule
Min. runtime (\si{\milli\second}) & 287.42 & 570.37 & 1137.39 & 1774.45\\
Max. runtime (\si{\milli\second}) & 287.83 & 570.68 & 1138.96 & 1777.48\\
Avg. runtime (\si{\milli\second}) & 287.67 & 570.49 & 1138.00 & 1775.22\\
Jitter (\si{\milli\second})       & 0.41   & 0.31   & 1.57   & 3.03   \\ 
Jitter (\si{\percent})            & 0.14   & 0.05   & 0.14   & 0.17   \\
\midrule
Avg. time / trajectory (\si{\milli\second}) & 17.98 & 17.83 & 17.78 & 17.75 \\
\bottomrule
\end{tabular}
\end{table}

\begin{figure}[!ht]
    \centering
    \tikzsetnextfilename{runtime_boxplot}%
    \begin{tikzpicture}[font=\footnotesize]
  \begin{axis}[
    ylabel shift={3pt},
    trim axis left,
    width=7.1cm,
    height=2.5cm,
    /pgf/number format/1000 sep={},
    scale only axis,
    scaled ticks=false,
    scaled ticks=false,
    xmin=1137,
    xmax=1140,
    boxplot/draw direction=x,
    xlabel={Runtime in \si{\milli\second}},
    ylabel={Run},
    ytick={1,2,3,4},
    yticklabels={1, 2, 3, 4},
    boxplot/box extend=0.8,
  ]
    \addplot+[
      mark=*,
      mark options={color=black, scale=0.7},
      fill=Blue,
      boxplot prepared={
        average=1137.862245,
        median=1137.797375,
        lower quartile=1137.652250,
        upper quartile=1138.010687,
        lower whisker=1137.479750,
        upper whisker=1138.426750,
      },
      draw=black
    ] coordinates {
    (1, 1138.745750)
    };
    \addplot+[
      mark=*,
      mark options={color=black, scale=0.7},
      fill=Bluelight,
      boxplot prepared={
        average=1137.792500,
        median=1137.796062,
        lower quartile=1137.660500,
        upper quartile=1137.904625,
        lower whisker=1137.295750,
        upper whisker=1138.257250,
      },
      draw=black
    ] coordinates {
    (2, 1138.839125)
    };
    \addplot+[
      mark=*,
      mark options={color=black, scale=0.7},
      fill=Gray,
      boxplot prepared={
        average=1138.089149,
        median=1137.827875,
        lower quartile=1137.651750,
        upper quartile=1138.727750,
        lower whisker=1137.480625,
        upper whisker=1139.269375,
      },
      draw=black
    ] coordinates {
    };
    \addplot+[
      mark=*,
      mark options={color=black, scale=0.7},
      fill=LightGray,
      boxplot prepared={
        average=1138.003188,
        median=1137.826125,
        lower quartile=1137.668250,
        upper quartile=1138.342250,
        lower whisker=1137.391750,
        upper whisker=1138.963750,
      },
      draw=black
    ] coordinates {
    };
  \end{axis}
\end{tikzpicture}%

    \vspace{-4mm}
    \caption{Measured runtime distribution of the embedded planner for four independent runs (500 iterations, each 64 trajectories). The results demonstrate highly deterministic execution with sub-millisecond jitter.}
    \label{fig:boxplot_board}%
\end{figure}
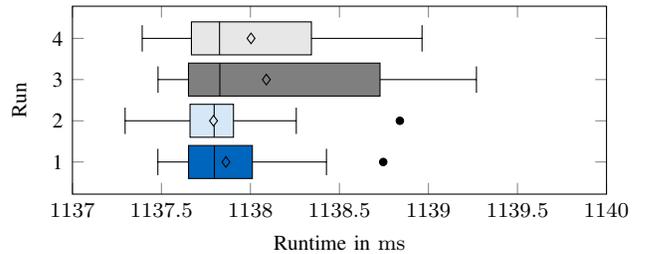

The total runtime increases linearly with the number of trajectories. The average computation time per trajectory remains nearly constant at around \SI{17.8}{\milli\second}, indicating that no additional overhead occurs as the workload increases. The observed jitter values are consistently around or below \SI{3}{\milli\second}, corresponding to less than \SI{0.2}{\percent} of the total runtime.
This high level of determinism is further supported by the results shown in Figure~\ref{fig:boxplot_board}.
The boxplot illustrates four independent runs, each comprising 500 planning cycles with 64 trajectories per cycle. Across all four runs, the execution time fluctuates only within a millisecond range, confirming that the RTOS provides reproducible timing over extended periods.

\begin{figure}[!ht]
\centering
    \tikzsetnextfilename{bar_chart_trajectory}%
    \newcommand{\plotheight}{2.7cm}
\newcommand{\plotwidth}{6.3cm}

\begin{tikzpicture}[font=\footnotesize]

\begin{axis}[
/pgf/number format/.cd,
1000 sep={},
xlabel shift={-2pt},
width  = \plotwidth,
height = \plotheight,
ybar stacked,
bar width=12pt,
scaled y ticks = false,
enlarge x limits=0.15,
axis y line*=left,
xlabel={Number of Samples},
ylabel={Latency in \si{\milli\second}},
ymin=0, ymax=2000,
xtick=data,
ymajorgrids=true,
major grid style={line width=.2pt,draw=gray!50},
scale only axis,
symbolic x coords={16,32,64,100},
legend style={
    at={(0.17,0.97)},
    anchor=north,
    legend columns=1, 
    column sep=0.5em,
    draw=none},
]

\addplot+[fill=Bluelight, draw=black] coordinates {
    (16,37.0) (32,74.3) (64,148.6) (100,232.1)
};

\addplot+[fill=Blue, draw=black, line width=0.4pt] coordinates {
    (16,246.2) (32,491.6) (64,984.9) (100,1538.6)
};

\legend{Generation, Evaluation}

\end{axis}


\begin{axis}[
/pgf/number format/.cd,
1000 sep={},
height=\plotheight,
width=\plotwidth,
enlarge x limits=0.15,
scale only axis,
scaled ticks=false,
scaled ticks=false,
ylabel={CPU utilization in \si{\percent}},
xmin=0,
xmax=1,
ymin=0, 
ymax=100,
axis x line*=none,
axis y line*=right,
ylabel shift={-4pt},
xticklabels={},
xtick={-10},
ytick={0, 25, 50, 75, 100},
]

\addplot [thick, Black, mark=*, mark size=1.5pt]
table {%
0.00 48 
0.33 67   
0.67 90   
1.00 99  
};

\end{axis}
\end{tikzpicture}%

\vspace{-8.0 mm}
\caption{Runtime breakdown and CPU utilization for different sample counts. The black line corresponds to the CPU utilization shown on the right y-axis.}
\label{fig:latency_cpu_combined}
\end{figure}
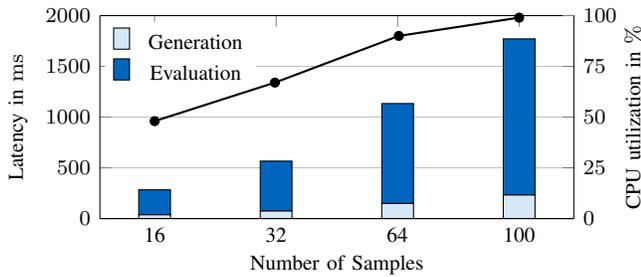

\autoref{fig:latency_cpu_combined} shows the breakdown of total runtime into trajectory generation and evaluation stages, along with the corresponding CPU utilization.
The evaluation accounts for the majority of the total computation time. This phase involves converting polynomial coefficients into discrete trajectory points, transforming coordinates from curvilinear to cartesian space, and evaluating feasibility and cost for all trajectories. These operations dominate the runtime and CPU usage, indicating that this stage provides great potential for optimization. While the absolute runtime is not yet sufficient for a primary motion planner, the achieved determinism and bounded latency make the proposed approach suitable as a backup or safety planner within a fail-operational system, as motivated in \autoref{fig:futureimpact}. In this use case, trajectories could be generated in the background and be activated when the main planner fails. Moreover, the results underline the importance of advanced sampling strategies. In the current implementation, samples are distributed uniformly across the action space, resulting in trajectory evaluations in irrelevant regions. Informed sampling approaches could concentrate computational effort on important subsets of the action space, for instance. This would reduce computational load without compromising the coverage of feasible trajectories.

\section{Conclusion \& Outlook}
\label{sec:conclusion}

This work presented the implementation and evaluation of a sampling-based motion planner executed on automotive-grade embedded hardware under real-time operating conditions. The planner was deployed on an Arm-based NXP S32Z2 platform running Zephyr \gls{rtos} and evaluated using the CommonRoad simulation framework. The results demonstrate that deterministic trajectory planning is feasible on constrained embedded systems, with constant execution times and minimal jitter. Although the absolute runtime is higher than that of an \gls{hpc}, the embedded solution shows better timing consistency and predictability. These characteristics are critical for safety-certified applications, where deterministic response behavior is more important than peak computational speed. Our presented concept, therefore, provides a promising foundation for a certified fallback planner that operates independently of the main \gls{ad} stack.

Future work will focus on transferring the concept into a comprehensive safety architecture, where the planner acts as a Minimal Risk Maneuver (MRM) module capable of generating trajectories in the event of a primary system failure. Furthermore, the concept will be validated on a real vehicle to evaluate its robustness under real-world conditions.


\bibliographystyle{IEEEtran}
\bibliography{literature}
\end{document}